\documentclass[smallabstract,smallcaptions]{dccpaper}
\usepackage{epsfig}
\usepackage{citesort}
\usepackage{amsmath}
\usepackage{amssymb}
\usepackage{color}
\usepackage{url}
\usepackage{amsmath}
\usepackage{amssymb}
\usepackage{fancyhdr}

\newlength{\figurewidth}
\newlength{\smallfigurewidth}

\begin{document}

\title
{\large
\textbf{Video Quality Assessment Based on Swin TransformerV2 and Coarse to Fine Strategy}
}

\author{%
Zihao Yu$^{\ast}$, Fengbin Guan$^{\ast}$, Yiting Lu$^{\ast}$, Xin Li$^{\ast}$, and Zhibo Chen$^{\ast}$\\[0.5em]
{\small\begin{minipage}{\linewidth}\begin{center}
\begin{tabular}{ccc}
$^{\ast}$University of Science and Technology of China \\
Hefei, Anhui, China\\
\url{{yuzihao,guanfb, luyt31415, lixin666}@ mail.ustc.edu.cn,} \\
\url{chenzhibo@ustc.edu.cn} \\
\end{tabular}
\end{center}\end{minipage}}
}

\maketitle
\thispagestyle{fancy}
\lhead{}
\lfoot{}
\cfoot{\small{\copyright~ 2024 IEEE.  Personal use of this material is permitted.  Permission from IEEE must be obtained for all other uses, in any current or future media, including reprinting/republishing this material for advertising or promotional purposes, creating new collective works, for resale or redistribution to servers or lists, or reuse of any copyrighted component of this work in other works.}}
\rfoot{}

\begin{abstract}
%The objective of reference-free video quality assessment is to assess the quality of the distorted video without reference to high-definition video. In this study, we introduce an enhanced spatial perception module pre-trained on multiple image quality assessment databases and a lightweight temporal fusion module to tackle the NR-VQA task. This model employs swin transformerv2 as a local-level spatial feature extractor to fuse these multi-stage features through a few transformer layers and utilizes a Temporal transformer for the spatiotemporal domain feature fusion of the video. To accommodate compressed videos of varying bit rates, we also incorporate a coarse-to-fine contrastive strategy to enhance the model’s feature discriminative capability for videos with different bit rates.
The objective of non-reference video quality assessment is to evaluate the quality of distorted video without access to reference high-definition references. In this study, we introduce an enhanced spatial perception module, pre-trained on multiple image quality assessment datasets, and a lightweight temporal fusion module to address the no-reference visual quality assessment (NR-VQA) task. This model implements Swin Transformer V2 as a local-level spatial feature extractor and fuses these multi-stage representations through a series of transformer layers. Furthermore, a temporal transformer is utilized for spatiotemporal feature fusion across the video. To accommodate compressed videos of varying bitrates, we incorporate a coarse-to-fine contrastive strategy to enrich the model’s capability to discriminate features from videos of different bitrates. This is an expanded version of the one-page abstract.
\end{abstract}

\Section{Introduction}

With the increase in video resolution, the necessity to process high-resolution video has become paramount. To address this issue, we employ Swin TransformerV2\cite{liu2022swin} for spatial perception. The advantage of Swin TransformerV2 lies in its ability to efficiently process large-scale data while maintaining computational efficiency. However, videos with different compression bitrates may pose challenges to the model’s distinguishable capabilities. \\
For compression distortion, we propose a coarse-to-fine contrastive strategy to equip the model with the ability to handle compressed videos of varying bitrates. From the global perspective, we introduce a group contrastive loss\cite{roy2023test}. This loss function empowers the model to coarsely discern differences in videos when processing different bitrates. However, group contrast loss may cause the model to lack sufficient discriminative ability when processing videos with the same bitrate. To mitigate this issue, from the local perspective,  we introduce rank loss. By employing rank loss, we can ensure that the model retains good discriminative ability when processing videos within the same bitrate.

\Section{Methodology}

\begin{figure}       %不带*单栏，带*双栏
	\centering
        \includegraphics[scale=0.4]{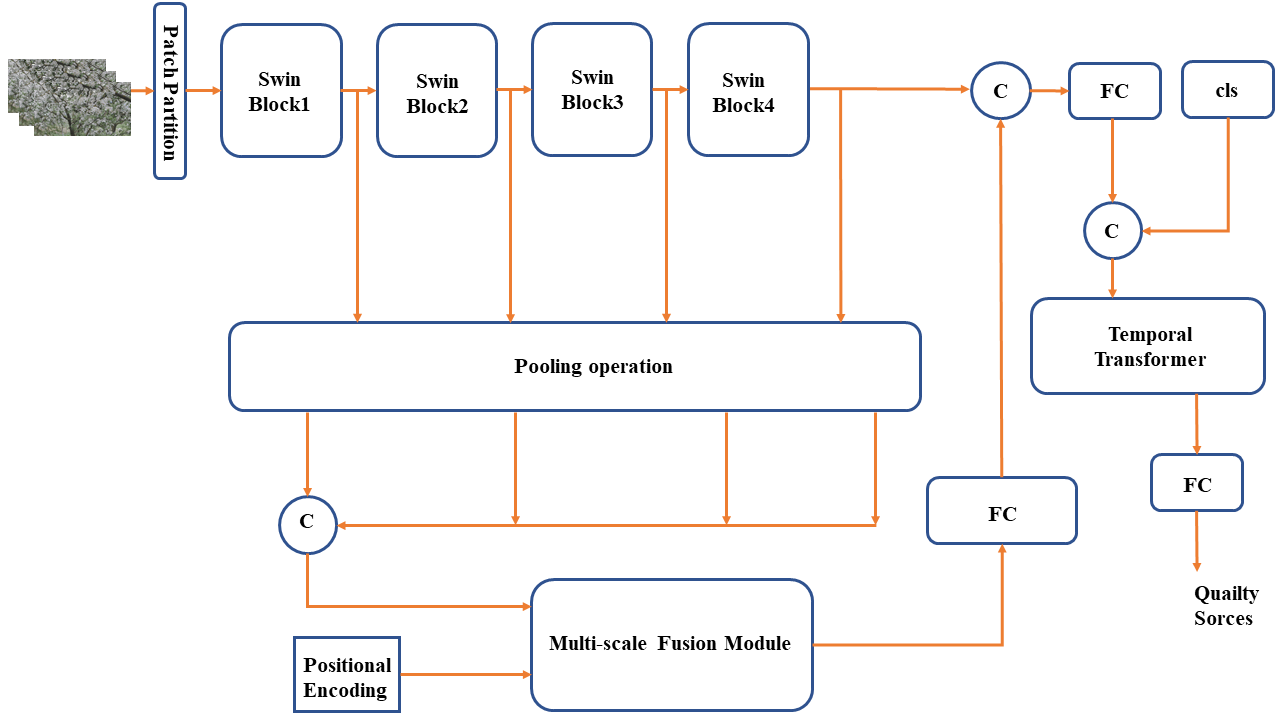}  %scale缩放比例，Fig.jpg文件名
	\caption{An overview of our proposed method}   % 图片名称
	\label{FIG:1}
\end{figure}

In this section, we delineate our proposed model as illustrated in Fig.~\ref{FIG:1}, a Non-reference Video Quality Assessment (NR-VQA) method comprising an enhanced spatial perception module and a lightweight temporal fusion module, equipped with coarse-to-fine contrastive strategy. \\

\SubSection{Feature Extraction}
In the spatial feature extraction part, we feed the sampled frames into SwinTransformerV2~\cite{liu2022swin}. SwinTransformerV2 is capable of obtaining local features of video frames at different levels. After normalizing and rescaling the local features through a feature pooling operation, we derive the global frame-level features by fusing the local representations across a series of transformer layers. Finally, we concatenate the local features and global features along the channel dimension and feed them into a fully connected layer to obtain the quality-aware features of the frames. This method fully utilizes global and local information, which helps to improve the perceptual ability of video quality assessment.\\
\SubSection{Coarse to Fine Contrastive Strategy}
In our model, we incorporate Group Contrastive loss as part of the loss function. We minimize the Group Contrastive loss that is defined as follows:\\
\begin{equation}
\mathcal{L}_{i, j}^{g c}=-\log \frac{\exp \left(\operatorname{sim}\left(\boldsymbol{z}_{(i)}, \boldsymbol{z}_{(j)}\right) / \tau\right)}{\sum_{k>(1-p) N}^N \exp \left(\operatorname{sim}\left(\boldsymbol{z}_{(i)}, \boldsymbol{z}_{(k)}\right) / \tau\right)}
\end{equation}
\begin{equation}
\mathcal{L}^{g c}=\sum_{i=1}^{p N} \sum_{\substack{j=1 \\ j \neq i}}^{p N} \mathcal{L}_{i, j}^{g c}+\sum_{i=(1-p) N+1}^N \sum_{\substack{j=(1-p) N+1 \\ j \neq i}}^N \mathcal{L}_{i, j}^{g c}
\end{equation}
Where $\tau$ is the temperature parameter, $p$ is a proportion parameter of $0.5$, $N$ is the size of the input vector , and $\text{sim}(z_i,z_j)$
is the similarity between vectors $z_i$ and $z_j$. 
%We build two dataloaders of the same number of videos of the two code rates and set $p$ to 0.5 so that the model can correctly process the videos of the two code rates.\\
The Group Contrastive loss minimizes the feature similarity within the same bitrate group and maximizes the feature similarity of the different bitrate groups. Through this approach, we can make the features of videos with the same bitrate closer, and the features of videos with different bitrates farther apart. This design enables our model to effectively handle multi-bitrate compressed video quality assessment.\\
We incorporate rank loss as part of the loss function to ensure that the model retains the discriminative ability when processing videos with the same bit rate. We minimize the rank loss that is defined as follows:\\
\begin{equation}
 \mathcal{L}^{Rank} = \sum_{i=1}^{n} \max(0, \text{margin} - \ (\text{pred}_1 - \text{pred}_2)_i (\text{label}_1 - \text{label}_2)_i) 
\end{equation}
Where $margin$ is a preset interval value, $pred$ is the predicted value, and label is the label value. The total loss for our model is defined as:\\
\begin{equation}
\mathcal{L} = \mathcal{L}^{MSE} + \mathcal{L}^{L1} + \lambda_1 \ \mathcal{L}^{g c} + \lambda_2 \ \mathcal{L}^{Rank} 
\end{equation}
\Section{Experiment}

\SubSection{Implementation details}
We pre-trained the spatial module on datasets such as CLIVE\cite{ghadiyaram2015massive}, LIVE\cite{sheikh2006statistical}, KonIQ-10k\cite{hosu2020koniq}, and KADID-10K\cite{lin2019kadid}. The diversity and extensiveness of these datasets provide our model with a wealth of training samples for enhancing image-level perception.\\
For performance evaluation, we employ two commonly used criteria, namely Spearman’s rank-order correlation coefficient (SROCC) and Pearson’s linear correlation coefficient (PLCC). Both SROCC and PLCC range from 0 to 1, and a higher value indicates a better performance.\\
During the pre-training phase, our spatial module was pre-trained using the loss function and training scheme of TReS\cite{golestaneh2022no}. TReS enhances the feature extraction capability of the model by using Relative Ranking and Self-Consistency. We trained on an NVIDIA 1080Ti GPU, with a batch size of 8, using an exponential decay scheme to adjust the learning rate, with a decay rate of 0.95.\\
\SubSection{Results}
As shown in Table~\ref{tab:example}, our spatial model achieved excellent results in the BIQA task. In our study, due to the lack of annotation for compressed video quality assessment, we decided to train and test on pseudo labels created using VMAF. This approach allows us to explore the performance of Video Quality Assessment (VQA) in an opinion-free manner.\\
As shown in Table~\ref{tab:example1}, our model achieved better performance on the pseudo-label dataset when using group contrastive loss and rank Loss. This indicates that the Coarse to Fine Contrastive Strategy is helpful for VQA tasks on compressed videos at different bit rates. This coarse-to-fine contrastive strategy facilitates perception for multi-bitrate video quality at both global and local level, empowering the model to gain an enhanced understanding and representation of changes in multiple distortion levels. This subsequently improves the predictive capabilities of the model.\\

\Section{Conclusion}
In this work, we propose an enhanced spatial perception module and lightweight temporal module implementing a coarse-to-fine contrastive strategy for compressed video quality assessment. For spatial feature extraction, we pre-train the spatial module on multiple blind image quality assessment (BIQA) datasets, achieving performance comparable to UNIQUE~\cite{zhang2021uncertainty} and TRES~\cite{golestaneh2022no}. This demonstrates the excellence of Swin Transformer V2 for image feature extraction. Concurrently, we hold high expectations for the coarse-to-fine contrastive strategy in the compressed video quality assessment across multiple bitrates. We believe this can further enrich the performance for compressed video quality assessment.\\

\begin{table}[tp]
\begin{center}
\caption{\label{tab:example}%
Performance comparison measured by SRCC and PLCC, where bold entries indicate the top one results.}
{
\renewcommand{\baselinestretch}{1}\footnotesize
\begin{tabular}{|c|c|c|c|c|c|c|c|c|}
\cline{1-9}
\multicolumn{1}{|c|}{~}&
\multicolumn{2}{c|}{LIVE} &
\multicolumn{2}{c|}{KADID} &
\multicolumn{2}{c|}{LIVEC} &
\multicolumn{2}{c|}{KonIQ}\\
\cline{2-9}
\multicolumn{1}{|c|}{\textbf{Method}} &
PLCC & SRCC & PLCC & SRCC & PLCC & SRCC & PLCC & SRCC\\
\hline
TReS\cite{golestaneh2022no}  &0.968 &0.969 &0.858 &0.859 &0.877 &0.846 &0.928 &0.915 \\
UNIQUE\cite{zhang2021uncertainty}  &0.968 &0.969 &0.876 &0.878 &0.890 &0.854 &0.901 &0.896 \\
MUSIQ\cite{ke2021musiq}  &0.911 &0.940 &0.872 &0.875 &0.746 &0.702 &0.928 &0.916 \\
DEIQT\cite{qin2023data}  &\textbf{0.982} &\textbf{0.980} &0.887 &0.889 &\textbf{0.894} &\textbf{0.875} &0.934 &0.921 \\
\hline
ours &0.965 &0.966 &\textbf{0.931} &\textbf{0.931} &0.883 &0.867 &\textbf{0.935} &\textbf{0.924} \\
\hline
\end{tabular}}
\end{center}
\end{table}

\begin{table}[tp]
\begin{center}
\caption{\label{tab:example1}%
Test Results of Coarse to Fine Contrastive Strategy}
{
\renewcommand{\baselinestretch}{1}\footnotesize
\begin{tabular}{|c|c|c|}
\hline
Method & PLCC & SRCC\\
\hline
GC and Rank & 0.924 & 0.903\\
\hline
no GC and Rank & 0.919 & 0.900\\
\hline
\end{tabular}}
\end{center}
\end{table}

\Section{References}
\bibliographystyle{IEEEbib}
\bibliography{refs}

\end{document}